%% file: root.tex
\pdfoutput=1 
\documentclass[letterpaper, 10 pt, conference]{ieeeconf}  

\IEEEoverridecommandlockouts                              

\usepackage{amsmath} 
\usepackage{amssymb}  
\usepackage{tabularx}
\usepackage{array,graphicx}
\usepackage[skip=3pt,font=small]{caption}
\usepackage{cite}

\input{includes.tex}

\title{\LARGE \bf
{BodySLAM++}: Fast and Tightly-Coupled\\Visual-Inertial Camera and Human Motion Tracking
}

\author{Dorian F. Henning$^{1}$, Christopher Choi$^{1}$, Simon Schaefer$^{2}$, Stefan Leutenegger$^{1,2}$
\thanks{This work was supported by Imperial College London, the Technical University of Munich (TUM), as well as the TUM AGENDA 2030, funded by the Federal Ministry of Education and Research (BMBF) and the Free State of Bavaria under the Excellence Strategy of the Federal Government and the L\"{a}nder as well as by the Hightech Agenda Bavaria.}
\thanks{$^{1}$Smart Robotics Lab, Dept.\ of Computing, Imperial College London, UK \newline {\tt\small \{d.henning,christopher.choi\}@imperial.ac.uk}}%
\thanks{$^{2}$Smart Robotics Lab, School of Computation, Information and Technology, Technical University of Munich, Germany {\tt\small firstname.surname@tum.de}}%
}

\begin{document}

\maketitle
\thispagestyle{empty}
\pagestyle{empty}

\begin{abstract}
\input{sections/0_abstract}

\end{abstract}

\section{INTRODUCTION}
\input{sections/1_introduction}

\section{RELATED WORK}
\input{sections/2_related_work}

\section{PRELIMINARIES}
\input{sections/3_preliminaries}

\section{SYSTEM OVERVIEW}
\input{sections/4_system_overview}

\section{FRONTEND OVERVIEW}
\input{sections/5_frontend}

\section{VISUAL-INERTIAL \& HUMAN ESTIMATOR}
\input{sections/6_vioh_estimator}

\section{EXPERIMENTAL RESULTS}
\input{sections/7_results}

\section{CONCLUSION}
\input{sections/8_conclusion}

\bibliographystyle{IEEEtran}
\bibliography{references}

\end{document}

%% file: includes.tex
\usepackage{xifthen}
\usepackage{xparse}
\usepackage{subdepth}
\usepackage{leftidx}
\usepackage{bm}

\newcommand{\bbm}{\begin{bmatrix}}
\newcommand{\ebm}{\end{bmatrix}}

\DeclareMathAlphabet{\mbf}{OT1}{ptm}{b}{n}
\newcommand{\mbs}[1]{{\bm{#1}}}

\newcommand{\mbsbar}[1]{{\overline{\boldsymbol{#1}}}}
\newcommand{\mbshat}[1]{{\hat{\boldsymbol{#1}}}}
\newcommand{\mbstilde}[1]{{\tilde{\boldsymbol{#1}}}}

\newcommand{\mbsdot}[1]{{\dot {\boldsymbol{#1}}}}

\newcommand{\mbfbar}[1]{{\overline{\mbf{#1}}}}
\newcommand{\mbfhat}[1]{{\hat{\mbf{#1}}}}
\newcommand{\mbftilde}[1]{{\tilde{\mbf{#1}}}}

\newcommand{\mbfdot}[1]{{\dot{\mbf{#1}}}}

\newcommand{\cframe}[1]{{\smash{\protect\underrightarrow{\mathcal{F}}_{#1}}}}

\DeclareMathAlphabet{\mathbfit}{OML}{cmm}{b}{it}
\newcommand{\homo}[1]{{\mathbfit{#1}}}

\newcommand{\mbfh}[1]{{\homo{#1}}}

\newcommand{\pos}[2]{\leftidx{_{#1}}{ \mbf r}{_{#2}}} 
\newcommand{\postilde}[2]{\leftidx{_{#1}}{\mbftilde r}{_{#2}}} 
\newcommand{\posh}[2]{\leftidx{_{#1}}{\mbfh r}{_{#2}}} 

\newcommand{\rotvec}[2]{\leftidx{_{#1}}{ \mbs \alpha }{_{#2}}}

\newcommand{\lmh}[1]{\leftidx{_{#1}}{\mbfh l}{}} 

\newcommand{\vel}[3]{\leftidx{_{#1}}{\mbf v}{\IfValueTF{#2}{_{#2#3\hspace{2pt}}}{}}} 
\newcommand{\veltilde}[3]{\leftidx{_{#1}}{\mbftilde v}{\IfValueTF{#2}{_{#2#3\hspace{2pt}}}{}}} 
\newcommand{\velbar}[3]{\leftidx{_{#1}}{\mbfbar v}{\IfValueTF{#2}{_{#2#3\hspace{2pt}}}{}}} 
\newcommand{\velhat}[3]{\leftidx{_{#1}}{\mbfhat v}{\IfValueTF{#2}{_{#2#3\hspace{2pt}}}{}}} 
\newcommand{\veldot}[3]{\leftidx{_{#1}}{\mbfdot v}{\IfValueTF{#2}{_{#2#3\hspace{2pt}}}{}}} 

\newcommand{\myvec}[2]{\leftidx{_{#2}\hspace{-1pt}}{\mbf #1}{}} 

\newcommand{\acc}[3]{\leftidx{_{#1}}{\mbf a}{\IfValueTF{#2}{_{#2#3\hspace{2pt}}}{}}} 
\newcommand{\acctilde}[3]{\leftidx{_{#1}}{\mbftilde a}{\IfValueTF{#2}{_{#2#3\hspace{2pt}}}{}}} 
\newcommand{\accbar}[3]{\leftidx{_{#1}}{\mbfbar a}{\IfValueTF{#2}{_{#2#3\hspace{2pt}}}{}}} 

\newcommand{\rotvel}[3]{\leftidx{_{#1}}{\mbs \omega}{\IfValueTF{#2}{_{#2#3\hspace{2pt}}}{}}} 
\newcommand{\rotveltilde}[3]{\leftidx{_{#1}}{\mbstilde \omega}{\IfValueTF{#2}{_{#2#3\hspace{2pt}}}{}}} 
\newcommand{\rotvelbar}[3]{\leftidx{_{#1}}{\mbsbar \omega}{\IfValueTF{#2}{_{#2#3\hspace{2pt}}}{}}} 
\newcommand{\rotvelhat}[3]{\leftidx{_{#1}}{\mbshat \omega}{\IfValueTF{#2}{_{#2#3\hspace{2pt}}}{}}} 
\newcommand{\rotveldot}[3]{\leftidx{_{#1}}{\mbsdot \omega}{\IfValueTF{#2}{_{#2#3\hspace{2pt}}}{}}} 

\newcommand{\C}[2]{\leftidx{}{\mbf C}{_{#1#2\hspace{2pt}}}} 
\newcommand{\T}[2]{\leftidx{}{\mbfh T}{_{#1#2\hspace{2pt}}}} 
\newcommand{\q}[2]{\leftidx{}{\mbf q}{_{#1#2\hspace{2pt}}}} 
\newcommand{\qtilde}[2]{\leftidx{}{\mbftilde q}{_{#1#2\hspace{2pt}}}} 

\newcommand{\real}{\mathbb{R}}
\newcommand{\Kset}{\mathcal{K}}
\newcommand{\Jset}{\mathcal{J}}
\newcommand{\Lset}{\mathcal{L}}
\newcommand{\Hset}{\mathcal{H}}
\newcommand{\Pset}{\mathcal{P}}
\newcommand{\Cset}{\mathcal{C}}
\newcommand{\et}{\emph{et al.} }

\newcolumntype{Y}{>{\centering\arraybackslash}X}
\newcolumntype{Z}{>{\raggedleft\arraybackslash}X}

\newcommand{\Log}[1]{\text{Log}\left( #1 \right)}

%% file: sections/0_abstract.tex
Robust, fast, and accurate human state -- 6D pose and posture -- estimation remains a challenging problem.
For real-world applications, the ability to estimate the human state in real-time is highly desirable.
In this paper, we present {BodySLAM++}, a fast, efficient, and accurate human and camera state estimation framework relying on visual-inertial data.
{BodySLAM++} extends an existing visual-inertial state estimation framework, OKVIS2, to solve the dual task of estimating camera and human states simultaneously.
Our system improves the accuracy of both human and camera state estimation with respect to baseline methods by 26\% and 12\%, respectively, and achieves real-time performance at 15+ frames per second on an Intel i7-model CPU.
Experiments were conducted on a custom dataset containing both ground truth human and camera poses collected with an indoor motion tracking system.

%% file: sections/1_introduction.tex
In applications such as human-robot cooperation, interaction, augmented and virtual reality, fast and accurate human shape, posture, and 6D pose estimation remains a challenging problem.
Such tasks rely on real-time estimation, and also benefit from a dense human representation, for example, a parametric human mesh, over skeleton-based representations.

State-of-the-art methods currently utilise powerful deep neural networks to estimate dense human representations.
However, these methods require high computational resources that are not always practical in the field.
Additionally, these methods often come with a caveat that large amounts of annotated data are required to train these models, which is both difficult and labour-intensive to obtain at a large scale.

\begin{figure}[ht!]
    \centering
    \includegraphics[width=\linewidth]{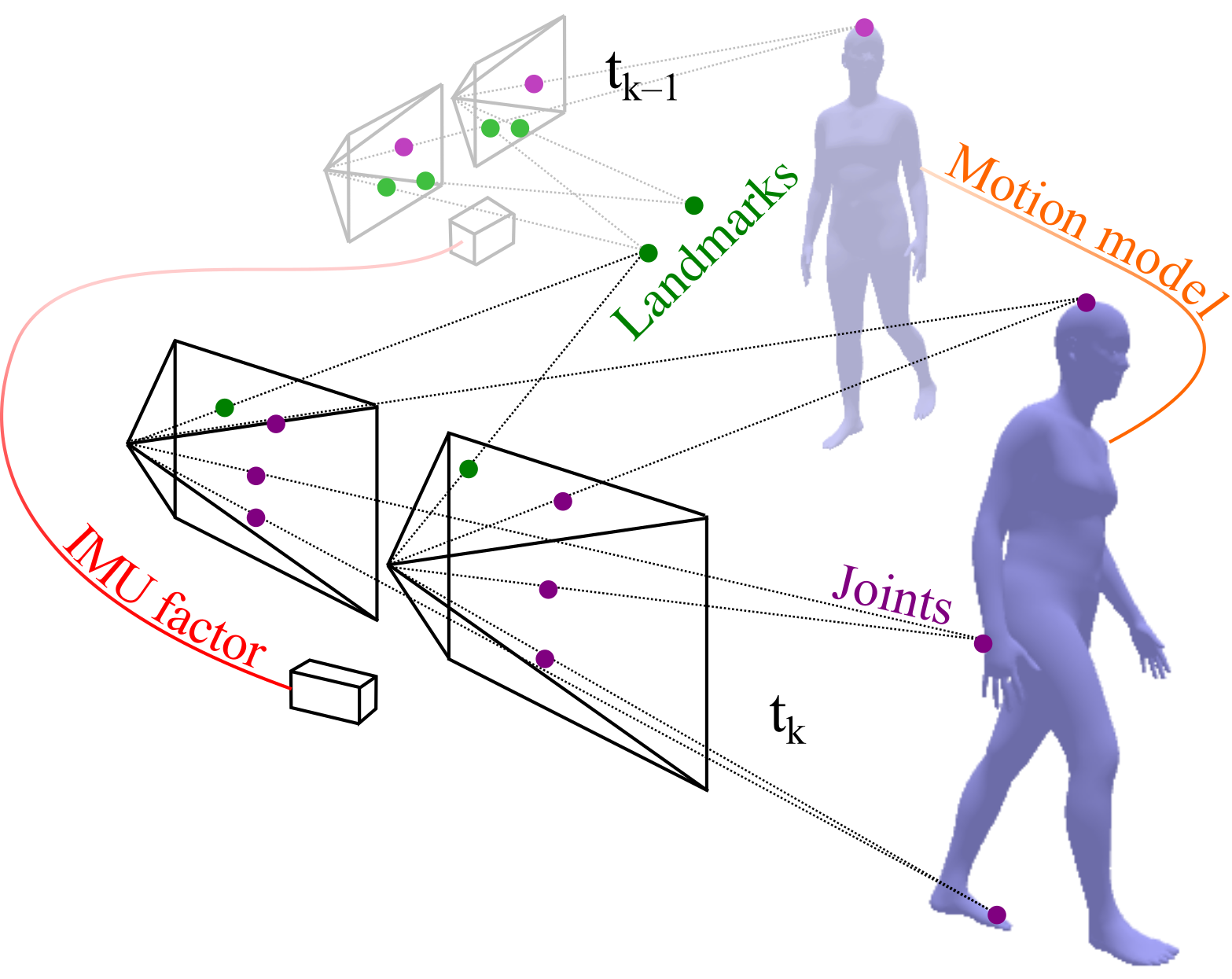}
    \caption{Conceptual overview of the {BodySLAM++} framework.
    We use an IMU factor between frames, as well as a human motion model to predict the displacement of the sensor and human, respectively.
    The human joints and static landmarks are reprojected into the individual camera frames where they are visible.}
    \label{fig:intro_figure}
\end{figure}

By leveraging a state-of-the-art Visual-Inertial state estimation framework, we can overcome the cumbersome reliance on expensive and indoor-only motion capture systems to provide globally consistent and causal human pose estimates.
These human mesh estimates can then be used for important robotic tasks, or to robustify and fine-tune advanced Computer Vision models on challenging ``in-the-wild'' data.

In this paper, we extend our previous work~\cite{Henning2022BodySLAMJointCamera}, where it was limited to a monocular camera and offline batch-processing, and propose a method called {BodySLAM++} that jointly estimates human posture, shape, and scale-aware (6D) poses of both humans and camera in real-time using a Visual-Inertial (VI) sensor in a tightly coupled manner that includes a learned human motion model.
We do this by extending OKVIS~2~\cite{Leutenegger2022OKVIS2RealtimeScalable}, a Visual-Inertial SLAM system, and used the SMPL parametric mesh model~\cite{Loper2015SMPLskinnedmulti} to represent the human body pose, shape and posture.
Additionally, we propose a novel human tracking technique, relying on the real-time estimation and propagation of the human states further leveraging our motion model.
To the best of our knowledge, this is the fastest, computationally least expensive parametric human mesh estimator that is tightly coupled to provide real-time output that can be used for robotic systems, and beyond.

To validate our proposed method, and to encourage more research on stereo human pose prediction, we collected and release an evaluation dataset.
Our proposed method, {BodySLAM++}, is evaluated on this dataset in a series of quantitative experiments.
The human pose estimation accuracy is increased as we can demonstrate a 26\% reduction in mean per joint error over the baseline method.
Furthermore, we also increase the sensor state estimation robustness in populated scenes, which is shown by a 12\% reduction of the average trajectory error.

In summary, our key contributions are:
\begin{itemize}
    \item{A factor-graph-based approach that estimates the human posture, shape, and 6D poses of both humans and VI-sensor in real-time.}
    \item{Improved accuracy of human and camera state estimation and tracking on recorded sequences with densely populated scenes.}
    \item{A dataset that we plan to release for benchmarking, which includes more than 8000 stereo camera frames, inertial sensor data, as well as ground truth camera poses and 22 human joints captured with an optical tracking system.}
\end{itemize}

%% file: sections/2_related_work.tex
\subsection{Visual-inertial Odometry / SLAM}
Visual-inertial Odometry / {SLAM} is a well-researched topic with state-of-the-art systems based on filtering~\cite{Mourikis2007MultiStateConstraint,Geneva2020OpenVINS,Bloesch2017IteratedextendedKalman} as well as factor-graphs and nonlinear least-squares~\cite{Qin2017VINSMono,Leutenegger2015Keyframebasedvisual,Campos2021ORBSLAM3,Leutenegger2022OKVIS2RealtimeScalable}, all estimating the state of the VI-sensor in a tightly coupled manner.
Please see e.g.~\cite{Leutenegger2022OKVIS2RealtimeScalable} for a comprehensive literature review. 
Our method builds on OKVIS~2~\cite{Leutenegger2022OKVIS2RealtimeScalable} and substantially extends it with tightly-coupled human state estimation.

\subsection{Human State Estimation in 3D}
In the following, we provide a brief overview of commonly used keypoint-based human 6D pose and posture estimation techniques, followed by 3D human mesh models, and finally mesh estimation techniques with a focus on optimisation-based approaches that are closely related to our method.

\subsubsection{Keypoint-based 3D Skeleton Estimation}
Keypoint-based human pose estimation relies on sparse, 2D keypoint detections that are either ``lifted'' to 3D or triangulated using multi-view geometry.
For a more in-depth review, Kulkarni \et recently published an extensive survey article \cite{Kulkarni2023PoseAnalyser}.
Here, only the directly related and relevant literature is presented.
Cao \et proposed a widely adopted 2D human skeleton estimation pipeline called {OpenPose}~\cite{Cao2017Realtimemultiperson}. 
This framework is robust in estimating multiple person through the use of affinity fields for the part association.
Other popular estimation frameworks with similar performance and adaption are AlphaPose \cite{Fang2017RMPERegionalMulti} or MaskRCNN \cite{He2017MaskRCNN}.
{OpenPose} was chosen as the human keypoint detection frontend, because of its unique API to interface with our optimisation-based SLAM framework, and for its status as a pseudo-ground-truth for 2D keypoint annotation.

\subsubsection{Human Representations}
There are different ways of representing human bodies, with skeleton and human parametric meshes being the most widely adopted ones.
The use of parametric human mesh models is advantageous compared to skeleton-based representations as they find a sweet spot in terms of complexity for volumetric human representation and still provide easily enforceable constraints on joint angles, shapes, and surface contact through their low-dimensional parameter space.
The most widely adapted and successful human mesh model is the SMPL (skinned multi-person linear) model \cite{Loper2015SMPLskinnedmulti} with the expressive extension SMPL-X \cite{Pavlakos2019ExpressiveBodyCapture}.
This is the model used in our formulation and further explained in \ref{sec:humanmeshmodel}.
A more recent model proposed by Osman \et is STAR~\cite{Osman2020STARSparseTrained}, which distinguishes itself mainly through the sparse representation of mesh correlations, and Quaternion-based rotation representation in 3D.
Our approach allows a quick substitution of the human mesh representation by an updated version like {STAR}, with minimal influence on the proposed optimisation factors but without loss of generality.

\subsubsection{Regression-based Mesh Model Prediction}
Many works have been published for single-image human mesh prediction, with impressive results \cite{Kolotouros2019LearningReconstruct3D,Kanazawa2018EndendRecovery,Kocabas2021SPECSeeingPeople,Li2021Hybrikhybridanalytical}.
These works lay the groundwork for many multi-image works on which we will focus.
The seminal work~\cite{Kanazawa2019Learning3DHuman} uses a feature-level temporal encoder to learn human dynamics from video input and can even hallucinate human motion from a single image input.
In VIBE~\cite{Kocabas2020VIBEVideoInference}, human dynamics were encoded using a recurrent architecture that was trained using a motion discriminator leveraging a large human motion database~\cite{Mahmood2019AMASSArchiveMotion}.
More recently, MEVA \cite{Luo20203DHumanMotionEstimation} was proposed, increasing the plausibility and accuracy of human motion prediction from videos by compressing and refining the motion using a variational autoencoder.
While all these works produce accurate root-relative joint positions from video inputs, they struggle to produce consistent motions in a global coordinate frame.
Approaches like GLAMR~\cite{Yuan2022GLAMRGlobalOcclusion} and {BodySLAM}~\cite{Henning2022BodySLAMJointCamera} achieve globally feasible motions from dynamic cameras; but those are non-causal batch-processing approaches, requiring information unavailable to online and real-time systems.
Our approach only relies on past information and is, to the best of our knowledge, the first attempt of a causal human mesh estimation framework.

\subsubsection{Optimisation-based Mesh Model Estimation}
Instead of using whole image end-to-end regression of human mesh parameters, one can directly fit human models to underlying observations like silhouettes or keypoints.
The early work {SMPLify}~\cite{Bogo2016KeepitSMPL} fits the {SMPL} model~\cite{Loper2015SMPLskinnedmulti} to 2D keypoints, optimising for the human mesh parameters.
A temporally consistent approach was proposed in~\cite{Arnab2019Exploitingtemporalcontext}, constraining the human mesh and motion between frames.
One limitation of optimisation-based approaches is unresolved ambiguities that come from fitting a high-dimensional parametric model to a few 2D keypoint measurements.
To resolve this shortcoming, one has to rely on a multitude of hand-engineered or trained priors, such as\cite{Pavlakos2019ExpressiveBodyCapture,Davydov2022AversarialParametricPose,Rempe2021HuMoR3DHuman}, or the more recently proposed neural distance field prior for human postures \cite{Tiwari2022PoseNDF}.
Our approach tightly integrates such human priors into our global optimisation factor graph such that we can leverage their power to not only increase the accuracy and plausibility of the human mesh estimation but also improve the camera state estimation.

%% file: sections/3_preliminaries.tex
\subsection{Notation}
We use the following notation throughout this work:
a reference coordinate frame is denoted as $\cframe{A}$, and position vectors expressed in this frame are written as $\pos{A}{}$, where as $\pos{A}{OP}$ denotes a translation vector from $O$ to $P$ expressed in $\cframe{A}$.
The homogeneous transformation from reference coordinate frame $\cframe{B}$ to $\cframe{A}$ is denoted as $\T{A}{B}$.
The rotation matrix of $\T{A}{B}$ is denoted as $\C{A}{B} \in SO(3)$, whose minimal representation is parameterised as an axis-angle rotation vector in the respective frame, $\rotvec{A}{AB} \in \real^3$.
Furthermore, the rotation matrix $\C{A}{B}$ can also be represented by a Hamilton Quaternion $\q{A}{B}$ that describes the attitude of $\cframe{B}$ relative to $\cframe{A}$.
Commonly used reference frames are the static world frame $\cframe{W}$, the IMU coordinate frame $\cframe{S}$, the camera frames $\cframe{C}_i, i = 1, ..., N$, and the human body centric frames $\cframe{H}_m, m = 1, ..., M$.
Camera frames are indexed by time steps $k$, and body joints by $j$.
Measurements of quantity $\mbf{z}$ are denoted with a tilde, $\mbftilde{z}$.

\subsection{Human Mesh Model}
\label{sec:humanmeshmodel}
The SMPL model \cite{Loper2015SMPLskinnedmulti,Pavlakos2019ExpressiveBodyCapture} is a parametric human mesh model that supplies a function, $\mathcal{M}(\mbs{\beta}, \mbs{\theta})$, of the shape parameters, $\mbs{\beta}$, and posture parameters, $\mbs{\theta}$, and returns a mesh, $_H\mbf{M} \in \mathbb{R}^{N \times 3}$, as a set of $N = 6890$ vertices in the human-centric frame $\cframe{H}$.

A single body shape is parameterised as a 10-dimensional vector, $\mbs{\beta}\in \mathbb{R}^{10}$, where the components are determined by a principal component analysis to capture as much of the variability in human body shape as possible.

The body posture parameters, $\mbs{\theta}$, represent each of the 23 joint orientations in the respective parent frame in axis-angle representation with dimensionality $\mathrm{dim}(\mbs{\theta}) = 23 \times 3 = 69$.
Other works \cite{Zhou2019ContinuityRotationRepresentations,Kolotouros2019LearningReconstruct3D} suggest the use of a 6-dimensional rotation representation, or parameterise orientation of the joints as Hamilton Quaternions \cite{Osman2020STARSparseTrained}.
However, since the human mesh parameters are optimised in our work, we prefer a minimal representation of rotations.
Furthermore, the Ceres Solver \cite{Agarwal2022CeresSolver} is capable of performing optimisation steps on the rotation Manifold $SO(3)$.

With this kinematic tree representation, it is easy to encode the same posture for different human shapes, as the joint positions can be defined as a linear combination of a set of mesh vertices.
For $J$ joints, a linear regressor $\mbf{W} \in \mathbb{R}^{J \times N}$ is pre-trained to regress the $J$ 3D body joints as $\myvec{X}{H} = \mbf{W} \: _H\mbf{M}$ with $\myvec{X}{H} \in \mathbb{R}^{J \times 3}$.

%% file: sections/4_system_overview.tex
\begin{figure}
    \centering
    \includegraphics[width=\linewidth]{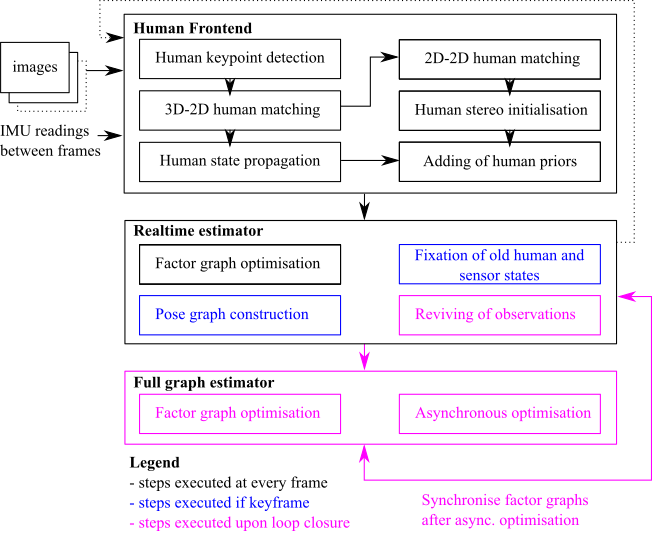}
    \caption{Overview of {BodySLAM++}: The frontend takes stereo camera images, detects humans, initialises them through triangulation, or matches them to existing states from the estimator. The real-time estimator will optimise current human states and fix older ones. Loop closures are optimised upon detection in the full graph estimator and synchronised with the realtime graph.}
    \label{fig:system_overview}
\end{figure}

The {BodySLAM++} system extends the \emph{frontend} and \emph{real-time estimator} of an underlying traditional SLAM system (i.e.\ {OKVIS~2}~\cite{Leutenegger2022OKVIS2RealtimeScalable}).
The system can be configured to account for loop closures.
Whenever a new (multi-) frame arrives, the images and IMU messages are processed synchronously with the SLAM system.
As shown in the system overview in \ref{fig:system_overview}, the \emph{human frontend} deals with human keypoint detection, human data association, human state initialisation, and human 3D tracking and state propagation.
The \emph{real-time estimator} will then optimise the human and camera states in the respective factor graph.
Posegraph edges are created from the \emph{real-time estimator} when marginalising old observations.
The human 6D poses and the corresponding motion factors are synchronised with the full graph, such that feasible human motions upon loop closures are enforced.

%% file: sections/5_frontend.tex
In the following, the human and SLAM frontend will be briefly discussed.

\subsection{SLAM Frontend}
The visual SLAM frontend is adopted from OKVIS 2~\cite{Leutenegger2022OKVIS2RealtimeScalable} and briefly summarised here. It extracts BRISK 2~\cite{Leutenegger2011BRISKBinaryRobust,Leutenegger2014UnmannedSolarAirplanes} keypoints and descriptors from every (multi-) frame image.
The 2D keypoints are then matched to already existing 3D landmarks in the map.
For the matching, both descriptor distance and 3D reprojected distance are considered.
The remaining unmatched keypoints are initialised from stereo triangulation.
Keyframe selection is discussed in \cite{Leutenegger2015Keyframebasedvisual}.

To eliminate non-static keypoints, the segmentation network Fast-SCNN~\cite{Poudel2019FastSCNNFS} trained on Cityscapes and run on all keyframe images.
Keypoints that represent slowly moving landmarks such as clouds, that are not ignored by RANSAC, are discarded in this step to improve outdoor accuracy.
Human keypoints are not removed by the segmentation network.

\subsection{Human Frontend}
The frontend for human detection and tracking consists of a human keypoint detector that runs synchronously to the SLAM frontend.
State-of-the-art keypoint detectors such as OpenPose \cite{Cao2017Realtimemultiperson} or AlphaPose \cite{Fang2017RMPERegionalMulti} are fast and resource-efficient enough that they run in real-time on the GPU.

To allow for frame-to-frame association of the human subjects, we use a 3D-2D tracking method.
All currently tracked humans that were visible in the previous few frames are propagated into the current frame with our human motion model described in Sec.~\ref{sec:motion_model}.
The propagated humans are then projected into the current frame and associated with 2D human keypoint detections via 2D overlap and root-aligned 2D mean joint distance. 

All 2D human keypoint detections that are not associated with existing propagated human states are then matched in the same manner to other, non-associated stereo detections, and used to initialise new humans via stereo triangulation.

%% file: sections/6_vioh_estimator.tex
In the following section, the extended factor graph from OKVIS~2~\cite{Leutenegger2022OKVIS2RealtimeScalable} and BodySLAM~\cite{Henning2022BodySLAMJointCamera} is explained.
Furthermore, the newly introduced factors that are unique to this visual-inertial and human SLAM formulation are introduced in detail.

\subsection{Overview and Optimisation Factor Graph}
\begin{figure}
    \centering
    \includegraphics[width=0.95\linewidth]{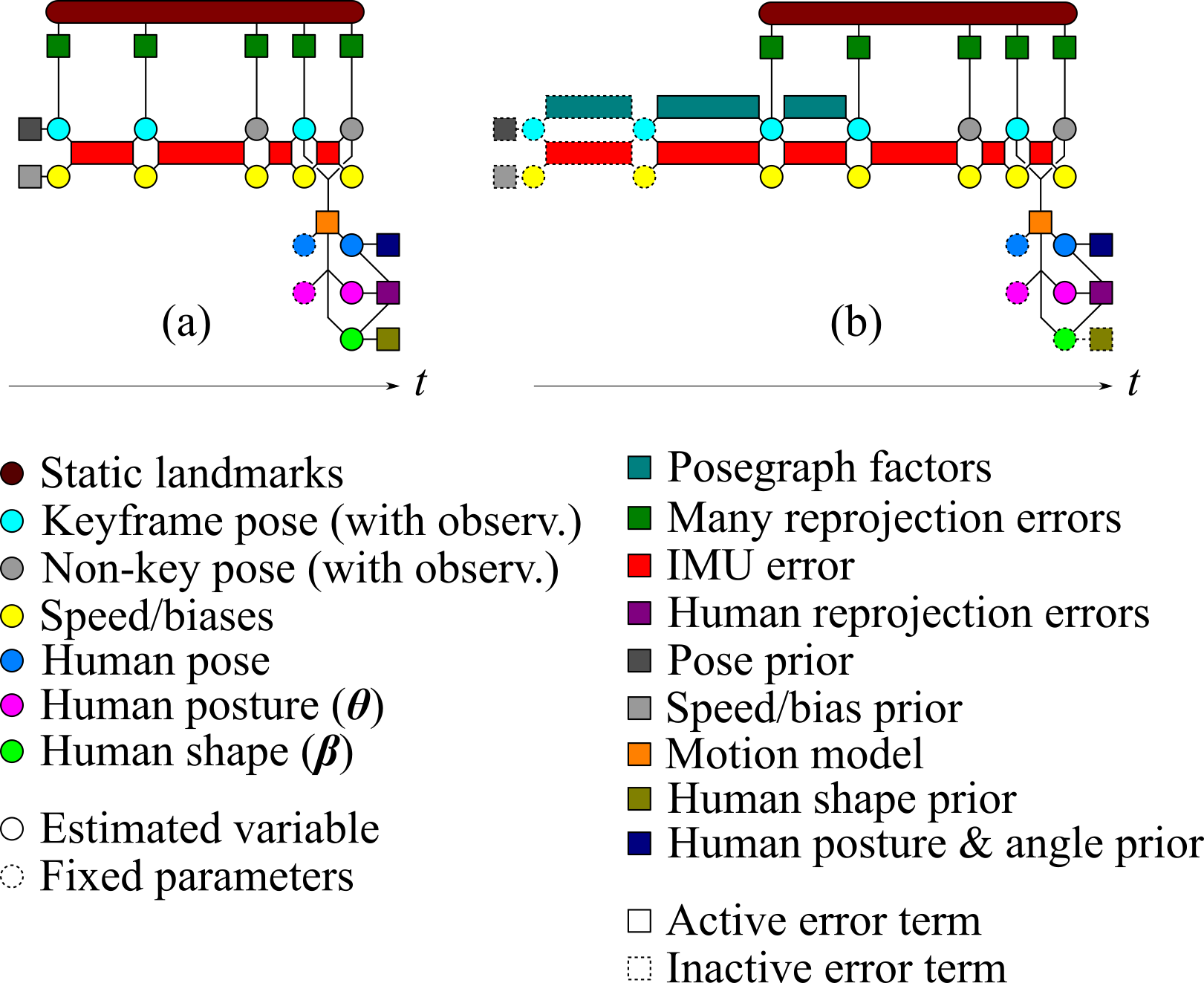}
    \caption{A full realtime estimator factor graph is created and shown in Fig.~\ref{fig:factor_graph}a. Motion model factors are tightly coupled to the posegraph, while the human reprojection errors and the linear priors are independent of the sensor poses. Later, keyframe poses are connected through relative pose errors as shown in Fig.~\ref{fig:factor_graph}b, similar to OKVIS~2~\cite{Leutenegger2022OKVIS2RealtimeScalable}. All human factors are set to be inactive after initial optimisation.}
    \label{fig:factor_graph}
\end{figure}

The optimisation factor graph is explained in Figure~\ref{fig:factor_graph}, whereby the individual factors, i.e.\ error terms forming the graph edges will be detailed below.
In comparison to OKVIS~2, we are adding frame-to-frame factors between human and camera poses, as well as anthropometric priors for the human posture, shape, and joints.

In the factor graph optimisation, we jointly estimate sensor state vector:
\begin{equation}
  \mbf{x}^\mathrm{sensor} = [\pos{W}{W S}^T, \q{W}{S}^T, \vel{W}{S}{}^T, \mbf{b}_g^T, \mbf{b}_a^T],
\end{equation}
\noindent
and a human state vector:
\begin{equation}
  \mbf{x}_h^\mathrm{human} = [\pos{S}{S H_h}^T, \q{S}{H_h}^T, \mbs{\theta}_h^T, \mbs{\beta}_h^T],
\end{equation}
\noindent
for each tracked human $h$.
The individual components of $\mbf{x}^\mathrm{sensor}$ are the position of the sensor origin $\pos{W}{W S}$ in the world frame $\cframe{W}$, the Hamilton Quaternion of orientation $\q{W}{S}$ describing the attitude of $\cframe{S}$ with respect to $\cframe{W}$, and the velocity $\vel{W}{S}{}$ of the sensor origin relative to to $\cframe{W}$.
We also include the gyroscope and accelerometer biases $\mbs{b}_g$ and $\mbs{b}_a$ in the estimation, respectively.
Analogous to the sensor state vector, the human state vector $\mbf{x}^\mathrm{human}$ is composed of the position $\pos{S}{S H_h}$ and Hamilton Quaternion of orientation $\q{S}{H_h}$ of the human root joint relative the the sensor frame $\cframe{S}$.
Furthermore we also include the frame-wise human posture vector $\mbs{\theta}_h$ and the human shape parameter $\mbs{\beta}_h$.
The states are estimated at each time step $k$ when new camera frames are obtained, except for the human shape parameter $\mbs{\beta}_h$, which is set constant after a few observations of the tracked human.

\subsection{Error Terms}
Here, we present the individual error terms of our optimisation factor graph, and their respective Jacobians.
Jacobians for linear priors are omitted, as they are constant.

\subsubsection{Human Joint Reprojection Errors}
The human reprojection error term is one of the key components of our factor graph formulation.
In contrast to BodySLAM, we parameterise the human 6D pose with respect to the sensor pose.
This difference allows us to treat the relative poses between human subjects and sensors constant in older frames (after optimisation), a valid choice even in case of loop closures:
\begin{equation}
  \mbf{e}_{i,h,k,j}^{\mathrm{joints}} = \mbftilde{z}_{i,h,k,j} - \mbf{u}
      \left(
          \T{S_k}{C_i}^{-1}\T{S_k}{H_{h,k}} \lmh{H}_{j}(\mbs \theta_{h,k}, \mbs \beta_h)
      \right),
\end{equation}
\noindent
where $\mbftilde{z}_{i,h,k,j}$ denotes the OpenPose keypoint measurement of the $j$-th joint of the $h$-th human in the $i$-th image at timestep $k$, and $\lmh{H}_{j}(\mbs \theta_{h,k}, \mbs \beta_h)$ denotes the SMPL joint $j$ of the $h$-th human in frame $k$ expressed as homogeneous points in the respective human centered frame $\cframe{H}$ as a function of the posture $\mbs \theta_{h,k}$ and shape $\mbs \beta_h$.
The individual human joint reprojection errors are weighted with ${\mbf{W}^{\mathrm{joint}}_{i,k,h,j} = \sigma_{i,k,h,j}^{-2} \mbf{I}_{2\times2}}$, with $\sigma_{i,k,h,j}$ being the OpenPose keypoint confidence for the respective keypoint detection.
The robustness of the reprojection error term is increased by using a Cauchy cost function, in line with \cite{Leutenegger2015Keyframebasedvisual,Leutenegger2022OKVIS2RealtimeScalable}.
The Jacobian of the human joint reprojection error term is analog to to the Jacobian of static landmarks found in ~\cite{Leutenegger2015Keyframebasedvisual, Leutenegger2022OKVIS2RealtimeScalable}, chained with the Jacobian of the {SMPL}  model~\cite{Loper2015SMPLskinnedmulti}, $\mbf{E}_{\mathrm{SMPL}}(\mbs{\beta}, \mbs{\theta})$, which can be derived using a symbolic math toolbox.

\subsubsection{Human Shape Prior}
Similar to \cite{Bogo2016KeepitSMPL}, we use a human shape prior to constrain the possible values.
The SMPL model uses the 10 principal components of the average human body shape to contain as much variance as needed.
Large deviations from those components are penalised:
\begin{equation}
    \mbf{e}_{h}^{\beta} = \mbstilde{\beta}_{h} - \mbs{\beta}_{h}.
\end{equation}

If the shape of the subject is unknown, $\mbstilde{\beta}_{h}$ is represented by a zero vector, representing the average human body shape in the first 10 principal components.
Otherwise, the measurement of the human body shape can be used as a prior.
As the first principal component, $\beta_1$, encodes primarily the height of the observed person, this composes a strong prior on the scale of the human.
To minimize the potential cause of inconsistency in the optimisation problem, the covariance of the measurement is approximated by a diagonal matrix ${\mbs{\Sigma}^{\beta} = \operatorname{diag}(\sigma_{\beta_1}, \dots, \sigma_{\beta_{10}})}$, with ${\sigma_{\beta_{2,\dots,9}} = 1}$, ${\sigma_{\beta_1} \gg 1}$.
This reduces the influence of the scale prior.
The respective error weight is the information matrix ${\mbf{W}^{\beta} = \lambda^{\beta} {\mbs{\Sigma}^{\beta}}^{-1}}$, with a weighting factor $\lambda_{\beta}$.

\subsubsection{Human Posture Prior}
Several different human posture priors have been proposed in literature \cite{Bogo2016KeepitSMPL,Pavlakos2019ExpressiveBodyCapture,Davydov2022AversarialParametricPose}.
In our work, we decided on a Gaussian Maximum Mixture Model prior consisting of $g = 8$ Gaussians proposed in \cite{Bogo2016KeepitSMPL}, as it has the lowest computational complexity involved:
\begin{equation}
    \mbf{e}_{h,k}^{\theta} =
        \mbs{\theta}_{h,k} - \mbs{\mu}^{\theta}_g, 
\end{equation}
\noindent
with $\mbs{\mu}^{\theta}_g$ the mean of the Gaussian Mixture $g$ with the smallest, weighted Malahanobis distance to the human posture $\mbs{\theta}_{h,k}$.
The error weight is the information matrix ${\mbf{W}^{\theta}_g = {\mbs{\Sigma}^{\theta}_g}^{-1}}$ of the respective Gaussian Mixture.

Additionally, we use a joint angle prior as proposed first in \cite{Bogo2016KeepitSMPL}, to avoid hyperflexion of elbow and knee joints:
\begin{equation}
    \mbf{e}_{h,k}^{\alpha} =
        [- \theta_{h,k}^{\mathrm{elbow}}, - \theta_{h,k}^{\mathrm{knee}}]^T \in \real^2,
\end{equation}
\noindent
with $\theta_{h,k}^{\mathrm{elbow}}$ and $\theta_{h,k}^{\mathrm{knee}}$ being the pitch component of the elbow and knee joint, respectively, and an error weight ${\mbf{W}^{\alpha} = \lambda^{\alpha} \mbf{I}_{2\times2}}$.

\subsubsection{Motion Model Factor}
\label{sec:motion_model}
The motion model introduced in {BodySLAM}~\cite{Henning2022BodySLAMJointCamera} constraints both the relative change of the 6D pose and the posture of the humans.
Please take note, that the human index $h$ is omitted for readability.

The expected human translation in the next camera camera frame $\postilde{H_{k\text{-}1}}{} := \pos{H_{k\text{-}1}}{H_{k\text{-}1} H_{k} }^{\mathrm{pred}}$ is used in the human motion error term:
\begin{equation}
  \mbf{e}_{k}^{\mathrm{M\text{-}p}} = \postilde{H_{k\text{-}1}}{} - \left[
    \T{S_{k\text{-}1}}{H_{k\text{-}1}}^{-1} \T{W}{S_{k\text{-}1}}^{-1} \T{W}{S_k} \posh{S_k}{S_k H_k} \right]_{1:3},
\end{equation}
\noindent
where $\C{S}{H_{k\text{-}1}}$ denotes orientation of the human pose in the sensor frame and $\C{W}{S_{k\text{-}1}}$ the orientation of the sensor frame.
The error weight is modeled as ${\mbf{W}^{\mathrm{M\text{-}p}} = \lambda^{\mathrm{M\text{-}p}} \mbf{I}_{3\times3}}$.
The Jacobian with respect to the reduced human state vector $[\pos{S}{H}^T, \q{S}{H}^T]$ is:
\begin{equation}
\resizebox{\linewidth}{!}{%
$\bbm
    \mbf{E}_{k\text{-}1}^{\text{human}} \\
    \mbf{E}_{k}^\text{human}
\ebm
   =
\C{S_{k\text{-}1}}{H_{k\text{-}1}}^T
\bbm
        - \mbf{I}_{3\times3} &
        \left[
            \T{W}{S_{k\text{-}1}}^{-1} \T{W}{S_k} \posh{S_k}{S_k H_k} - \posh{S_{k\text{-}1}}{S_{k\text{-}1} H_{k\text{-}1}}
        \right]_{1:3}^{\times} \\
        \C{W}{S_{k\text{-}1}}^T \C{W}{S_{k}} &
        \mbf{0}_{3\times3},
\ebm$}.
\rule[-1.5em]{0pt}{0pt}
\end{equation}

The Jacobian with respect to the reduced sensor state vector $[\pos{W}{S}^T, \q{W}{S}^T]$ is:
\begin{equation}
\resizebox{\linewidth}{!}{%
$\bbm
    \mbf{E}_{k\text{-}1}^{\text{sensor}} \\
    \mbf{E}_{k}^\text{sensor}
\ebm
   =
\C{S_{k\text{-}1}}{H_{k\text{-}1}}^T \C{W}{S_{k\text{-}1}}^T
\bbm
        - \mbf{I}_{3\times3} &
        - \left[
            \T{W}{S_k} \posh{S_k}{S_k H_k} - \posh{W}{W S_{k\text{-}1}}
        \right]_{1:3}^{\times} \\
        - \mbf{I}_{3\times3} &
        \left[
            \C{W}{S_k} \pos{S_k}{S_k H_k}
        \right]^{\times}
\ebm$}.
\rule[-1.5em]{0pt}{0pt}
\end{equation}

Change in human posture is formulated as a linear error:
\begin{equation}
  \mbf{e}_{k}^{\mathrm{M\text{-}\theta}} = \Delta\mbstilde{\theta}_{k,k\text{-}1} - ( \mbs{\theta}_{k} - \mbs{\theta}_{k\text{-}1} ),
\end{equation}
\noindent
where $\Delta\mbstilde{\theta}_{k,k\text{-}1}$ is the expected change in posture predicted by our motion model.
The error weight of the posture change is ${\mbf{W}^{\mathrm{M\text{-}\theta}} = \lambda^{\mathrm{M\text{-}\theta}} \mbf{I}_{69\times69}}$.

\subsubsection{OKVIS~2 Error Terms}
We adopt standard reprojection error, IMU error and relative pose error from OKVIS~2.

The projection error $\mbf{e}_{i,k,l}^{\mathrm{lm}}$ of the $l$-th landmark $\lmh{W}_{l}$ into the $i$-th camera frame at timestep $k$:
\begin{equation}
    \mbf{e}_{i,k,l}^{\mathrm{lm}} = \mbftilde{z}_{i,k,l} - \mbf{u}(\T{S_k}{C_{i}}^{-1}\T{S_k}{W} \lmh{W}_{l}),
\end{equation}
\noindent
with $\mbftilde{z}_{i,k,l}$ being the keypoint detection and $\mbf{u}(.)$ denoting the projection function of a pinhole camera, with optional distortion models (equidistant or radial-tangential).

The IMU error $\mbf{e}_{k,n}^{\mathrm{s}}$ between time step $k$ and $n$ is:
\begin{equation}
    \mbf{e}_{k,n}^{\mathrm{s}} = \mbfhat{x}_{n}^{\mathrm{sensor}} ( \mbf{x}_{k}^{\mathrm{sensor}}, \mbftilde{z}_{k,n} ) \boxminus \mbf{x}_{n}^{\mathrm{sensor}},
\end{equation}
\noindent
with $\mbfhat{x}_{n}^{\mathrm{sensor}}$ denoting the predicted sensor state at time step $n$ based on the estimated sensor state $\mbf{x}_{k}^{\mathrm{sensor}}$ and the IMU readings $\mbftilde{z}_{k,n}$ between those frames.
The $\boxminus$ denotes a regular subtraction except for the Hamilton Quaternion: ${\q{}{} \boxminus \q{}{}^\prime = \Log{\q{}{} \otimes \q{}{}^\prime}}$.

The relative pose error $\mbf{e}_{r,c}^{\mathrm{p}}$ between time step $r$ and $c$ is:
\begin{equation}
    \mbf{e}_{r,c}^{\mathrm{p}} = 
    \mbf{e}^{\mathrm{p},0}_{r,c} +
    \begin{bmatrix}
        \pos{S_r}{S_c} - \postilde{S_r}{S_c} \\
        \q{S_r}{S_c} \boxminus \qtilde{S_r}{S_c}
    \end{bmatrix},
\end{equation}
\noindent
where $\postilde{S_r}{S_c}$, $\qtilde{S_r}{S_c}$ are nominal relative position and orientation respectively.
The constant $\mbf{e}^{\mathrm{p},0}_{r,c}$, any related Jacobians, and the error weights $\mbf{W}^{\mathrm{lm}}_{i,k,l}$, $\mbf{W}_{k}^{\mathrm{s}}$, and $\mbf{W}_{r,c}^{\mathrm{p}}$ for the OKVIS~2 error terms are discussed in detail in~\cite{Leutenegger2015Keyframebasedvisual, Leutenegger2022OKVIS2RealtimeScalable}.

\subsection{Real-time Estimation Problem}
The real-time human and camera pose estimation problem is minimising the following non-linear least squares cost:

\begin{align}
  c(\mbf{x}) 
    &= \frac{1}{2} \sum_{i} \sum_{k \in \Kset} \sum_{l \in \Lset(i,k)} \rho \left( {\mbf{e}_{i,k,l}^{\mathrm{lm}}}^{\mathrm{T}} \mbf{W}^{\mathrm{lm}}_{i,k,l} \mbf{e}_{i,k,l}^{\mathrm{lm}} \right) \nonumber \\
    &+ \frac{1}{2} \sum_{k \in \Pset \cup \Kset \setminus f} {\mbf{e}_{k}^{\mathrm{s}}}^{\mathrm{T}} \mbf{W}_{k}^{\mathrm{s}} \mbf{e}_{k}^{\mathrm{s}} + \frac{1}{2} \sum_{r \in \Pset} \sum_{c \in \Cset(r)} {\mbf{e}_{r,c}^{\mathrm{p}}}^{\mathrm{T}} \mbf{W}_{r,c}^{\mathrm{p}} \mbf{e}_{r,c}^{\mathrm{p}} \nonumber \\
    &+ \frac{1}{2} \sum_{h \in \Hset} \left[ \sum_{i} \sum_{k \in \Kset} \sum_{j \in \Jset} \rho \left( {\mbf{e}_{i,h,k,j}^{\mathrm{joints}}}^{\mathrm{T}} \mbf{W}^{\mathrm{joints}}_{i,h,k,j} \mbf{e}_{i,h,k,j}^{\mathrm{joints}} \right) \right. \nonumber \\ 
    &+ \sum_{k \in \Kset} {\mbf{e}_{h,k}^{\mathrm{M\text{-}\theta}}}^{\mathrm{T}} \mbf{W}^{\mathrm{M\text{-}\theta}} \mbf{e}_{h,k}^{\mathrm{M\text{-}\theta}} + \sum_{k \in \Kset} {\mbf{e}_{h,k}^{\mathrm{M\text{-}p}}}^{\mathrm{T}} \mbf{W}^{\mathrm{M\text{-}p}} \mbf{e}_{h,k}^{\mathrm{M\text{-}p}} \nonumber \\
    & \left. + \hspace{-0.5ex} \sum_{k \in \Kset} {\mbf{e}_{h,k}^{\theta}}^{\mathrm{T}} \mbf{W}^{\theta}_g \mbf{e}_{h,k}^{\theta} + \hspace{-0.5ex} \sum_{k \in \Kset} {\mbf{e}_{h,k}^{\alpha}}^{\mathrm{T}} \mbf{W}^{\alpha} \mbf{e}_{h,k}^{\alpha} + {\mbf{e}^{\beta}_h}^{\mathrm{T}} \mbf{W}^{\beta} \mbf{e}^{\beta}_h \right]
.
\end{align}

The set $\Kset$ contains the $T$ most recent frames as well as $M$ keyframes, $\Hset$ contains all observed and tracked humans, $\Lset$ contains the observed landmarks in all $\Kset$ frames, $\Jset$ is the set of OpenPose~\cite{Cao2017Realtimemultiperson} joints, $\Pset$ contains all posegraph frames, i.e., poses connected through relative pose errors, with the subset $\Cset(r) \subset \Pset$ that contains all poses connected to posegraph frame $r$, and $f$ denoting the current frame.

As every human is parameterised with different postures and 6D poses per observed frame, but only one body shape, we are adding the shape prior only once for each human, whereas all other anthropometric priors and keypoint-based errors are added for each frame.

%% file: sections/7_results.tex
\begin{figure*}[!ht]
    \centering
    \includegraphics[width=\textwidth]{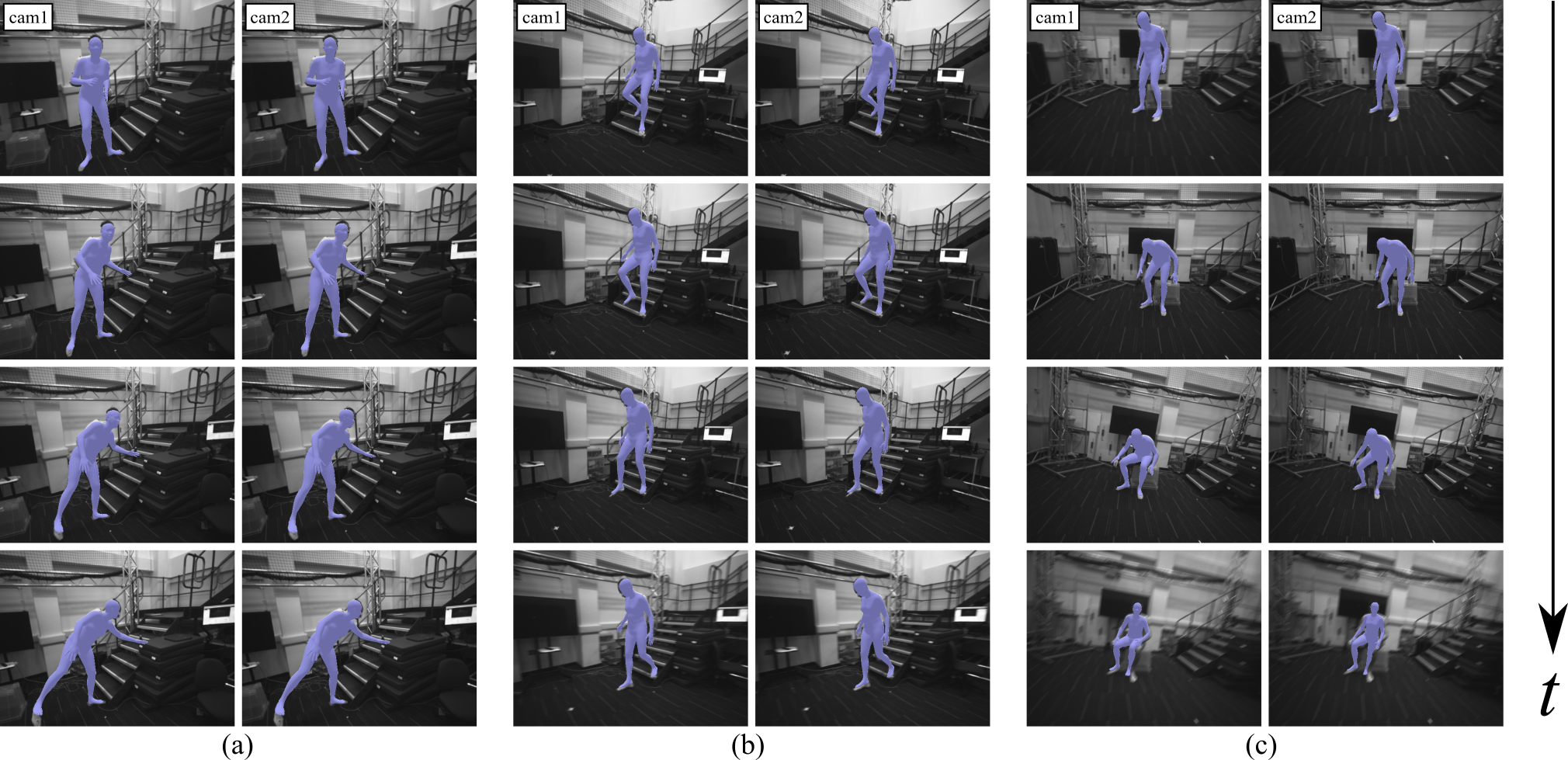}
    \caption{Qualitative results of our method {BodySLAM++}.
    We show the two camera frames \texttt{cam1} and \texttt{cam2}, over multiple time steps.
    Our system shows to be robust against our-of-distribution postures (a), uncommon camera angles (b), and severe motion blur (c).
    Furthermore, we demonstrate the temporal consistency of our method.}
    \label{fig:qualitative_results}
\end{figure*}

We evaluate the human 6D pose accuracy of our \mbox{BodySLAM++} framework in Sec.~\ref{sec:human_estimation} and the camera state estimation in Sec.~\ref{sec:camera_estimation}.
Real-time performance is shown in Sec.~\ref{sec:performance} and qualitative results are presented in Fig.~\ref{fig:qualitative_results}.

\subsection{Implementation Details}
\label{ref:impl_details}

\begin{table}[!b]
\caption{Mean Per Joint Position Error (MPJPE) of the single-person dataset in [m].
  The three baseline comparisons are a combination of a frame-wise SMPLify~\cite{Bogo2016KeepitSMPL} optimisation routine for the human shape, posture and 6D pose, and either ground truth, ORB-SLAM3 \cite{Campos2021ORBSLAM3} or OKVIS~2~\cite{Leutenegger2022OKVIS2RealtimeScalable} for the camera state estimation.
  Our method shows significant improvement across most sequences.
  In the first column, we furthermore compare against a regression-based approach, combining the results of a state-of-the-art human mesh regressor~\cite{Kolotouros2019LearningReconstruct3D} with the method proposed in~\cite{Henning2020HPE3D} to compute the relative camera to human coordinate transform.}
\label{tab:mpjpe_comparison}
\begin{tabularx}{\columnwidth}{l|Y|YYY|Y}
                & \rotatebox[origin=lb]{90}{\setlength\fboxsep{1cm} \parbox[c]{2.2cm}{\textbf{GT \\+ HPE3D}~\cite{Henning2020HPE3D}}}
                & \rotatebox[origin=lb]{90}{\setlength\fboxsep{1cm} \parbox[c]{2.2cm}{\textbf{GT}\\\textbf{+ SMPLify}~\cite{Bogo2016KeepitSMPL}}}
                & \rotatebox[origin=lb]{90}{\setlength\fboxsep{1cm} \parbox[c]{2.2cm}{\textbf{ORB-SLAM3}~\cite{Campos2021ORBSLAM3}\\\textbf{+ SMPLify}~\cite{Bogo2016KeepitSMPL}}}
                & \rotatebox[origin=lb]{90}{\setlength\fboxsep{1cm} \parbox[c]{2.2cm}{\textbf{OKVIS~2}~\cite{Leutenegger2022OKVIS2RealtimeScalable} \\\textbf{+ SMPLify}~\cite{Bogo2016KeepitSMPL}}}
                & \rotatebox[origin=lb]{90}{\setlength\fboxsep{1cm} \parbox[c]{2.2cm}{\textbf{{BodySLAM++} \\(ours)}}} \\ \hline
\texttt{\textbf{SP\_01}} & 0.8133   & 0.7675 & 0.7697          & 0.7672          & \textbf{0.2106} \\
\texttt{\textbf{SP\_02}} & 0.7865   & 0.6841 & 0.2059          & 0.2351          & \textbf{0.2041} \\
\texttt{\textbf{SP\_03}} & 0.1680   & 0.1258 & 0.1249          & 0.1216          & \textbf{0.1199} \\
\texttt{\textbf{SP\_04}} & 0.2248   & 0.2016 & 0.4395          & 0.1525          & \textbf{0.1391} \\
\texttt{\textbf{SP\_05}} & 0.2086   & 0.1846 & 0.2292          & \textbf{0.1453} & 0.1470          \\
\texttt{\textbf{SP\_06}} & -$^1$    & 0.1460 & 0.1824          & 0.1398          & \textbf{0.1354} \\
\texttt{\textbf{SP\_07}} & 0.1913   & 0.1613 & 0.1492          & 0.1830          & \textbf{0.1437} \\
\texttt{\textbf{SP\_08}} & 0.1985   & 0.1576 & 0.2746          & 0.1442          & \textbf{0.1335} \\
\texttt{\textbf{SP\_09}} & 0.1779   & 0.1508 & 0.1891          & \textbf{0.1240} & 0.1264          \\
\texttt{\textbf{SP\_10}} & 0.1573   & 0.1463 & 0.1425          & 0.1403          & \textbf{0.1361} \\
\texttt{\textbf{SP\_11}} & 0.1571   & 0.1562 & 0.1448          & 0.1494          & \textbf{0.1358} \\
\texttt{\textbf{SP\_12}} & 0.1817   & 0.2006 & 0.1828          & 0.1925          & \textbf{0.1504} \\
\texttt{\textbf{SP\_13}} & -$^1$    & 0.1610 & \textbf{0.1320} & 0.1399          & 0.1340          \\
\texttt{\textbf{SP\_14}} & -$^1$    & 0.2162 & 0.2520          & 0.1949          & \textbf{0.1202} \\
\texttt{\textbf{SP\_15}} & 0.1800   & 0.1416 & 0.1520          & 0.1382          & \textbf{0.1349} \\ \hline
\textbf{Avg.}            & (0.2871) & 0.2401 & 0.2380          & 0.1979          & \textbf{0.1447} \\ \hline
\multicolumn{6}{l}{\rule{0pt}{4ex}\footnotesize{
$^1$ HPE3D~\cite{Henning2020HPE3D} failed on those sequences.}}
\end{tabularx}
\end{table}

All experiments were performed on an Intel i7-12700K CPU with 3.4 GHz and 64 GB RAM. To achieve real-time performance we used Google's nonlinear least squares optimisation framework, Ceres-Solver~\cite{Agarwal2022CeresSolver}, and implemented the analytical Jacobians to help reduce computational requirements.
For evaluation purposes, the human keypoints were previously extracted using OpenPose \cite{Cao2017Realtimemultiperson} and stored as JSON, to minimize the influence of CPU to GPU synchronisation on the performance benchmarks.

We used the individual error term weights $\lambda^{\beta} = 0.1$, $\lambda^{\mathrm{M\text{-}p}} = 0.1$, $\lambda^{\mathrm{M\text{-}\theta}} = 3.0$, and $\lambda^{\alpha} = 0.4$.
Those values were determined experimentally.
Static landmark reprojections and IMU error terms are weighted in line with OKVIS~2~\cite{Leutenegger2022OKVIS2RealtimeScalable}.

The human body shape parameter is only introduced once for each human.
All reprojection and anthropometric priors are dependent on this factor.
This implementation is akin to having a body shape parameter per observation of a human, but it significantly reduces the computational complexity.

\begin{table}[!b]
\caption{Average Trajectory Error (ATE) on the multi-person dataset in [m].
  Our {BodySLAM++} system outperforms the two baseline methods in all but two sequences.}
\label{tab:ate_multi_person}
\begin{tabularx}{\columnwidth}{l|YYY}
                           & \mbox{\textbf{ORB-SLAM3}~\cite{Campos2021ORBSLAM3}} & \textbf{OKVIS~2}~\cite{Leutenegger2022OKVIS2RealtimeScalable} & \textbf{{BodySLAM++}} \\ \hline
\texttt{\textbf{MP\_01}}   & \textbf{0.0294}    & 0.0302          & 0.0337            \\
\texttt{\textbf{MP\_02}}   & 0.0305             & \textbf{0.0237} & 0.0250            \\
\texttt{\textbf{MP\_03}}   & 0.0364             & 0.0301          & \textbf{0.0257}   \\
\texttt{\textbf{MP\_04}}   & 0.0380             & 0.0476          & \textbf{0.0281}   \\
\texttt{\textbf{MP\_05}}   & 0.0271             & 0.0295          & \textbf{0.0249}   \\
\texttt{\textbf{MP\_06}}   & 0.0452             & 0.0350          & \textbf{0.0253}   \\
\texttt{\textbf{MP\_07}}   & 0.0306             & 0.0299          & \textbf{0.0276}   \\
\texttt{\textbf{MP\_08}}   & -$^1$              & 0.0277          & \textbf{0.0271}   \\
\texttt{\textbf{MP\_09}}   & -$^1$              & 0.0424          & \textbf{0.0374}   \\
\texttt{\textbf{MP\_10}}   & 0.1118             & 0.0581          & \textbf{0.0516}   \\
\texttt{\textbf{MP\_11}}   & 0.0618             & 0.0369          & \textbf{0.0329}   \\
\texttt{\textbf{MP\_12}}   & 0.0591             & 0.0385          & \textbf{0.0355}   \\
\texttt{\textbf{MP\_13}}   & 0.0667             & 0.0464          & \textbf{0.0412}   \\
\texttt{\textbf{MP\_15}}   & 0.0344             & 0.0303          & \textbf{0.0281}   \\ \hline
\textbf{Avg.}              & (0.0476)           & 0.0362          & \textbf{0.0317}   \\ \hline
\multicolumn{4}{l}{\rule{0pt}{4ex}\footnotesize{
$^1$ ORB-SLAM3~\cite{Campos2021ORBSLAM3} failed on those sequences.}}
\end{tabularx}
\end{table}

\subsection{Our Dataset}
\label{sec:dataset}
As part of our research contribution, we release our custom dataset for benchmarking.
The {BodySLAM} dataset consists of stereo visual-inertial data from a moving camera showing a single moving or stationary human, with associated ground truth information of the 6D camera pose and the 3D position of 22 human joints, tracked with an optical motion capture system (sequences \texttt{\textbf{SP\_01}} to \texttt{\textbf{SP\_15}}).
The captured motions cover typical slow to medium speed human activities such as walking, climbing and descending stairs, sitting down and standing up from a chair, and grabbing and placing objects.

\subsection{Comparison to baseline methods}
\label{sec:human_estimation}
In Table~\ref{tab:mpjpe_comparison}, we present the main results of our full \mbox{BodySLAM++} system.
As a baseline comparison, we estimate human shape, posture, and the 6D relative position between camera and human body with a frame-wise {SMPLify}-like optimisation routine~\cite{Bogo2016KeepitSMPL}, leaving out our contributed motion model factors, but using stereo joint observations where available.
To obtain a camera trajectory, we either use the ground truth (GT) from the optical tracking system, or perform classic visual-inertial SLAM on our data presented in \ref{sec:dataset} with state-of-the-art frameworks such as ORB-SLAM3~\cite{Campos2021ORBSLAM3} and OKVIS~2~\cite{Leutenegger2022OKVIS2RealtimeScalable}.
The mean per joint position error (MPJPE) of the tracked human is then computed with respect to the ground truth joint location by the external tracking system.
To allow a fair comparison, we aligned the full human root trajectories in 6D beforehand.
One can clearly see that our proposed system {BodySLAM++} outperforms the baseline methods with {SMPLify} in most of the sequences.

\subsection{SLAM improvement over baselines without human tracks}
\label{sec:camera_estimation}
Table~\ref{tab:ate_multi_person} presents the camera trajectory accuracy in comparison to state-of-the-art VI-SLAM systems {ORB-SLAM3} and {OKVIS~2} in presence of three humans in our proprietary multi-person dataset (sequences \texttt{\textbf{MP\_01}} to \texttt{\textbf{MP\_15}}).

\begin{figure}[!hb]
    \centering
    \includegraphics[width=\linewidth]{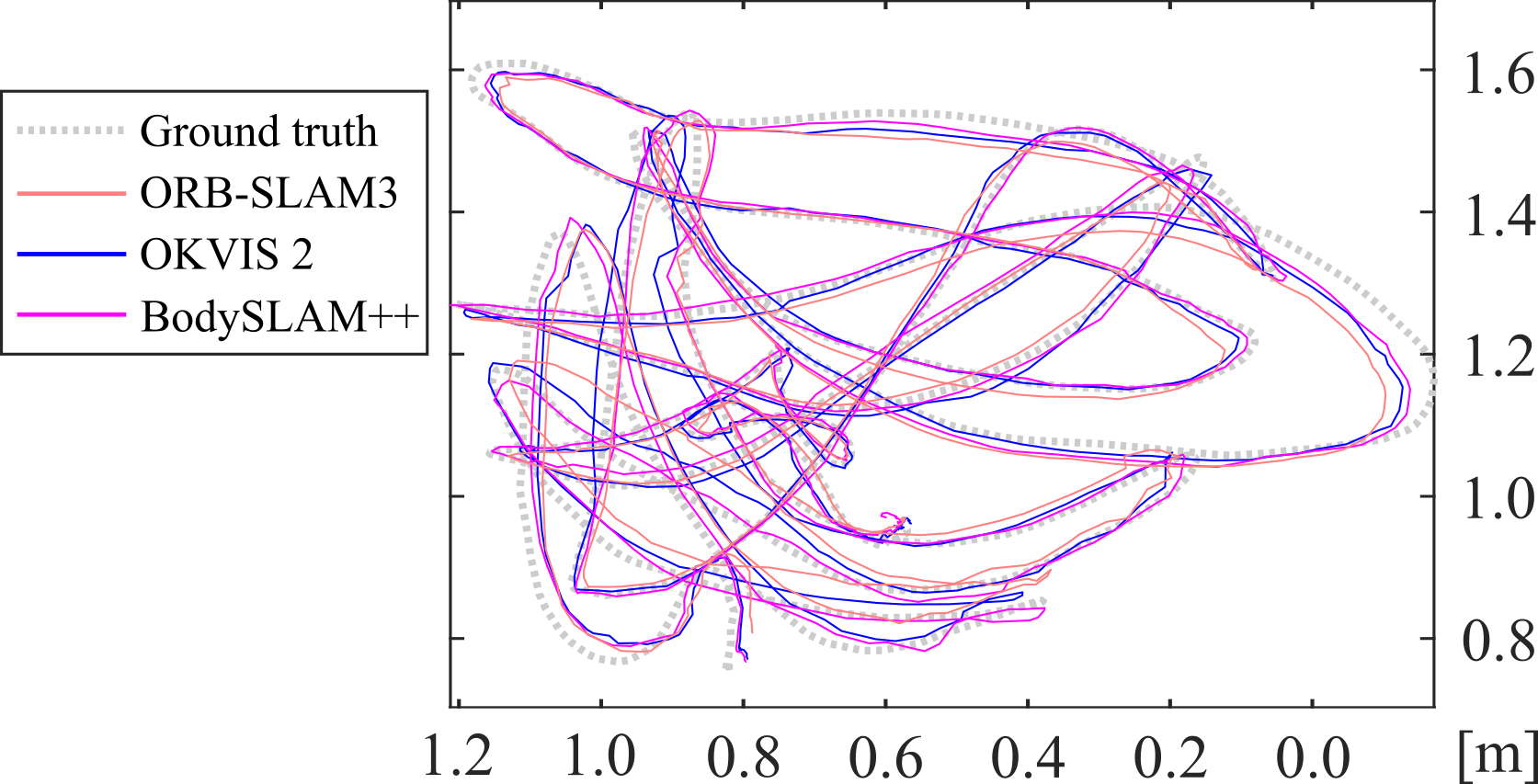}
    \caption{Trajectory plots on \texttt{MP\_03} sequence.
    {BodySLAM++} has the lowest Average Trajectory Error (ATE) with $0.0256$ m, compared to OKVIS~2 with $0.0301$ m, and ORB-SLAM3 with $0.0376$ m, showing superior robustness amongst people.}
    \label{fig:trajectory_plots}
\end{figure}

Except for two sequences, {BodySLAM++} outperforms the baseline methods on all sequences.
This shows that under extreme conditions, in scenes where multiple humans are occluding a large portion of tracked visual features, our contribution enables the optimisation to converge on a more accurate solution than without.
It appears that human reprojection errors in combination with the human motion model increase the robustness and contribute to higher accuracy and lower trajectory errors.
Furthermore, the {BodySLAM++} system additionally supplies accurately estimated human mesh models without compromising real-time capability.

In Fig.~\ref{fig:trajectory_plots}, the superior performance of the {BodySLAM++} framework is qualitatively showcased in comparison to the baseline OKVIS~2~\cite{Leutenegger2022OKVIS2RealtimeScalable} and ORB-SLAM3~\cite{Campos2021ORBSLAM3}.
It is discernible that over time, OKVIS~2 drifts further from the ground truth trajectory compared to our contribution, {BodySLAM++}.
This can be attributed to the increased accumulation of noisy keypoint measurements from the dynamic human body surface.

\begin{figure}[!hb]
    \centering
    \includegraphics[width=0.95\linewidth]{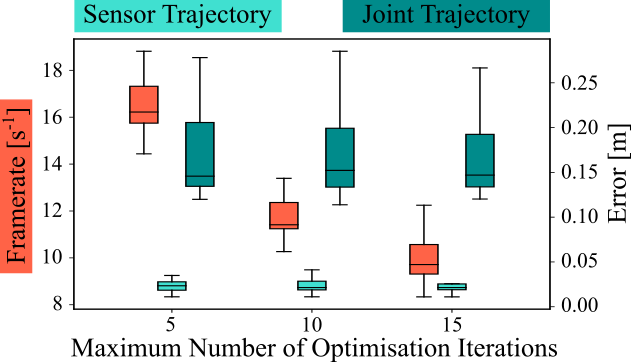}
    \caption{Performance of {BodySLAM++}: Frame-rate, average trajectory error (ATE), and mean per joint position error (MPJPE) for different maximum numbers of optimisation iterations. The error of both trajectory and human joints only slightly increases when constraining the number of iterations to achieve real-time performance.}
    \label{fig:timing_plots}
\end{figure}

\addtolength{\textheight}{-0.9cm}

\subsection{Real-time Performance}
\label{sec:performance}
To assess the real-time capability and corresponding accuracy of our proposed system, an experiment was performed where the real-time estimator was constrained to 5, 10, and 15 optimisation iterations.
The results are reported in Fig.~\ref{fig:timing_plots} and show the achieved frame-rate, and the two error metrics for the sensor (ATE) and joints (MPJPE) trajectories, respectively.

Constraining the maximum number of optimisation iterations significantly improves performance, up to more than the required framerate (\textgreater 15 FPS).
Furthermore, very little or no accuracy is sacrificed in doing so.
This effect can be explained by a marginal gain that is achieved by running the optimisation for the human mesh parameters longer than minimally required to fit the 2D keypoint detections.

%% file: sections/8_conclusion.tex
Real-time accurate human shape, posture, and 6D pose estimation is a challenging problem.
Deep Learning approaches require large amounts of annotated 3D data that are difficult to obtain.
In this paper, we extend our previous work, {BodySLAM}~\cite{Henning2022BodySLAMJointCamera}, and propose {BodySLAM++}, a factor-graph optimisation approach that jointly optimises camera poses, static landmarks, and human mesh parameters in a real-time fashion.
To the best of our knowledge, we are the first in using VI data to jointly estimate a dense human mesh model and metric trajectory from a dynamic VI-sensor.

To validate our method, we collected datasets containing visual-inertial data from a dynamic VI-sensor, as well as ground truth camera poses and human joint locations tracked via an indoor optical motion capture system, and demonstrated our proposed method has lower ATE and MPJPE compared to {ORB-SLAM3}~\cite{Campos2021ORBSLAM3} and {OKVIS~2}~\cite{Leutenegger2022OKVIS2RealtimeScalable}.

In future work, we would like to explore human posture priors to improve the estimation accuracy, and online sensor calibration to make system adoption easier.